\documentclass[11pt]{article}

\usepackage[margin=1in]{geometry}

\usepackage[T1]{fontenc}
\usepackage[utf8]{inputenc}

\usepackage{amsmath,amssymb,amsthm}

\usepackage{graphicx}%
\usepackage{multirow}%
\usepackage{mathrsfs}%
\usepackage[title]{appendix}%
\usepackage{xcolor}%
\usepackage{textcomp}%
\usepackage{manyfoot}%
\usepackage{booktabs}%
\usepackage{algorithm}%
\usepackage{algorithmicx}%
\usepackage{algpseudocode}%
\usepackage{listings}%

\usepackage[colorlinks=true,%
            linkcolor=blue,%
            citecolor=blue,%
            urlcolor=blue]{hyperref}

\newcommand{\bzeta}{{\boldsymbol \zeta}}

\newcommand{\blambda}{{\boldsymbol \lambda}}

\def\by{{\boldsymbol y}}
\def\bA{{\boldsymbol A}}
\def\bD{{\boldsymbol D}}
\def\bG{{\boldsymbol G}}
\def\bx{{\boldsymbol x}}

\def\bob{{\boldsymbol b}}
\def\bd{{\boldsymbol d}}
\def\bg{{\boldsymbol g}}
\def\bc{{\boldsymbol c}}
\def\bz{{\boldsymbol z}}
\def\bw{{\boldsymbol w}}

\def\bn{{\boldsymbol n}}
\def\b0{{\boldsymbol 0}}
\def\btau{{\boldsymbol \tau}}
\def\bxi{{\boldsymbol \xi}}

\def\bbR{{\mathbb R}}
\def\bbI{{\mathbb I}}
\def\bbZ{{\mathbb Z}}

\def\cG{{\mathcal G}}
\def\cD{{\mathcal D}}
\def\cL{{\mathcal L}}
\def\cK{{\mathcal K}}

\newtheorem{theorem}{Theorem}

\newtheorem{problem}[theorem]{Problem}

\title{Classification by Separating Hypersurfaces:\\
An Entropic Approach}

\author{%
  Argimiro Arratia\thanks{These authors contributed equally to this work. Department of Computer Science, Universitat Polit\`ecnica de Catalunya, Jordi Girona s/n, Barcelona 08034, Spain. Email: \texttt{argimiro.arratia@upc.edu}}%
  \and 
  Mahmoud El Daou\footnotemark[1]\thanks{Department of Computer Science, Universitat Polit\`ecnica de Catalunya, Jordi Girona s/n, Barcelona 08034, Spain. Email: \texttt{mahmoud.el.daou@upc.edu}}%
  \and 
  Henryk Gzyl\thanks{Center for Finance, IESA School of Business, Caracas 1010, Venezuela. Email: \texttt{henryk.gzyl@iesa.edu.ve}}%
}

\date{}

\begin{document}

\maketitle

\begin{abstract}
We consider the following classification problem: Given a population of individuals characterized by a set of attributes represented as a vector in $\bbR^N$, the goal is to find a hyperplane in $\bbR^N$ that separates two sets of points corresponding to two distinct classes. This problem, with a history dating back to the perceptron model, remains central to machine learning.
In this paper we propose a novel approach by searching for a vector of parameters in a bounded $N$-dimensional hypercube centered at the origin and a positive vector in $\bbR^M$, obtained through the minimization of an entropy-based function defined over the space of unknown variables. The method extends to polynomial surfaces, allowing the separation of data points by more complex decision boundaries. This provides a robust alternative to traditional linear or quadratic optimization techniques, such as support vector machines and gradient descent. Numerical experiments demonstrate the efficiency and versatility of the method in handling diverse classification tasks, including linear and non-linear separability.
\end{abstract}



\section{Introduction and preliminaries}
To state the problem in its simplest form, we follow \cite{BV}. Let $\cG=\{\bg_1,\ldots,\bg_{K_1}\}$ and $\cD=\{\bd_1,\ldots,\bd_{K_2}\}$ be two disjoint subsets, to be interpreted as describing the characteristics of two samples of ``good'' and ``defective'' individuals. 
In its simplest statement the classification problem of interest is stated as follows:

\begin{problem}\label{P1}
Given subsets $\cG=\{\bg_1,\ldots,\bg_{K_1}\}$ and $\cD=\{\bd_1,\ldots,\bd_{K_2}\}$ of $\bbR^N,$ determine a $\bw\in\bbR^N$ and $b_1,b_2>0$ such that
\begin{gather}
\langle\bg_i,\bw\rangle - b_1 > 0,\;\;i=1,\ldots,K_1. \label{prob1.1}\\
\langle\bd_i,\bw\rangle + b_2  < 0,\;\; i=1,\ldots,K_2 \label{prob1.2}
\end{gather}
\end{problem}

Writing $b_1=b+t$ and $b_2=b-t$ we recover the simple (for $t=0$) and robust (when $t>0$) statements of the linear discrimination problem. The term perceptron refers to a class of classification algorithms, of which the linear discrimination is but a small part. Its origin can be traced back to \cite{McP}, \cite{Ro} and \cite{Bo}. A few references to the subject in the physics literature are \cite{RN,Ki,Wa,NS}. Some mathematical aspects are dealt with in \cite{CZ} and \cite{SP}. For connections to pattern classification and statistical machine learning see \cite{DHS,MRT,CFO,HG}.

In these references, different variations on the theme of \eqref{prob1.1}-\eqref{prob1.2} are considered. Here we consider yet another statement of the same problem.

\begin{problem}\label{P2}
Given subsets $\cG=\{\bg_1,\ldots,\bg_{K_1}\}$ and $\cD=\{\bd_1,\ldots,\bd_{K_2}\}$ of $\bbR^N,$ determine a $\bw\in\bbR^N$ and $\bob_1\in\bbR^{K_1}_{++}$, $ \bob_2\in\bbR^{K_2}_{++}$ such that:
\begin{gather}
\langle\bg_i,\bw\rangle - b_1(i) = 0,\;\;i=1,\ldots,K_1. \label{prob2.1}\\
\langle\bd_i,\bw\rangle + b_2(i) = 0,\;\; i=1,\ldots,K_2 \label{prob2.2}
\end{gather}
\end{problem}

Note that the inequalities have now become equalities, and that the components of $\bob_1$ and $\bob_2$ are required to be positive. The geometric intuition behind this proposal is that, instead of aiming at separating planes from the outset, we search for a sheaf of parallel planes, passing through each point of the training data set $\cG\bigcup\cD$. After that we order them according to their distance to the hyperplane $\{\bz\in\bbR^N|\langle\bw,\bz\rangle=0\}$, in units of $\|\bw\|$. This ordering will determine three classification domains. The points that belong to two of them will be good or defective with certainty, and the points in the third region are classified with an empirically determined probability, that is, with uncertainty. This will be explained in detail below at the end of Section \ref{sec:4}.

To vectorize problem \ref{P2} put $M=K_1+K_2$ and consider the following $M\times N$ matrix $\bD$ whose first $K_1$ rows are $\bg_i$, $i=1,\ldots,K_1$,  and the next $K_2$ rows are $-\bd_i$, $i=1,\ldots,K_2.$ Now set
\begin{equation}\label{not1}
\bA = [\bD\, ,-\bbI_M],\;\; \bx={\bw^t\atopwithdelims()\bob},\;\;\bob={\bob_1^t\atopwithdelims()\bob_2^t}.
\end{equation}
Here the superscript ``$t$'' denotes transposition, $\bbI_M$ stands for the identity matrix in $\bbR^M$, $[A,B]$ denotes the concatenation of matrices $A$ and $B$, therefore $\bA$ is an $M\times(N+M)$-matrix and $\bx$ an $N+M$-vector whose components are required to be in the constraint set $\cK=[-1,1]^N\times\bbR^M_{++}.$ With these notations, our problem becomes:

\begin{problem}\label{P3}
Determine $\bx\in\cK$ such that
\begin{equation}\label{prob3}
\bA\bx = \b0.
\end{equation}
\end{problem}

Notice that we constrain $\bw$ to be in a box. This constraint is easier to deal with numerically than requiring $\|\bw\|^2\leq 1$.

The approach to solve problem \ref{P1} in \cite{BV,CZ,SP,DHS,MRT,CFO} and \cite{HG} involves variations of a constrained quadratic optimization procedure. An approach based on the maximum entropy method in the mean (see \cite{GG}) was proposed in \cite{GHM}. In that approach a problem similar to \ref{P1} is transformed into a problem of determining a probability measure on $\cK,$ which is then solved by the entropy maximization method as in \cite{J} or \cite{Ka}. The approach we propose here consists of minimizing a strictly convex function $\Psi(\bx)$ defined on $\cK,$ that looks like an entropy, but it is not an entropy since it is not defined on a class of probabilities to begin with, and its domain $\cK$ contains nonpositive vectors.  The optimization problem of interest is then the following:

\begin{problem}\label{P4}
Find $\bx^*\in\cK$ such that
\begin{equation}\label{prob4}
\bx^* = \arg\min\{\Psi(\bx)\mid \bx\in\cK,\;\bA\bx=\b0\}.
\end{equation}
\end{problem}

Not only is the objective function $\Psi$ not a quadratic function, but the constraints are equality constraints, and the solution yields the orientation of the separating plane and the distances of the training data to the separating plane.

The remainder of the paper is organized as follows. In Section \ref{sec:2}, we show how to extend the separation model to allow nonlinear classification using polynomial surfaces. In Section \ref{sec:3} we define the strictly convex function $\Psi:\cK \to \bbR$ and study some of its properties. In Section \ref{sec:4} we solve problem \ref{P4}. In Section \ref{sec:experiments} we compare the competitive computational efficiency and good performance of our Entropic Classifier (both linear and polynomial) against traditional classification methods, such as Support Vector Machines (SVM), K-Nearest Neighbors (KNN), Logistic Regression, and Perceptron, in different contexts.  

\section{Separation by polynomial surfaces}\label{sec:2}
Sometimes the data may suggest that the two training sets are separated by a polynomial surface. To be precise, for $\bx=(x_1,\ldots,x_N)\in\bbR^N$  and $\bn=(n_1,\ldots,n_N)\in\bbZ_{+}^N,$ let us write $\bx^{\bn}=\prod_{i=1}^Nx_i^{n_i}$ and refer to it as a monomial of degree $\sum_i n_i,$ and write $|\bn|=\sum n_i.$ Then a polynomial of degree $p$ can be written as:
\begin{equation}\label{poly1}
f(\bx) = \sum_{|\bn|\leq p} c(\bn)\bx^{\bn} = \sum_{k=0}^p \sum_{|\bn|=k} c(\bn)\bx^{\bn}.
\end{equation}

The polynomial version of problem \eqref{P1} can be stated as:
\begin{problem}\label{P5}
Given two sets $\cG=\{\bg_1,\ldots,\bg_{K_1}\}$ and $\cD=\{\bd_1,\ldots,\bd_{K_2}\}$ of $\bbR^N$ of training data, determine a polynomial $f(\bx)$ of degree $p$ such that:
\begin{gather}
f(\bg_i) - b_1 > 0,\;\;i=1,\ldots,K_1. \label{prob5.1}\\
f(\bd_i)+ b_2  < 0,\;\; i=1,\ldots,K_2. \label{prob5.2}
\end{gather}
\end{problem}

Notice that when the polynomial is homogeneous of degree $1$ we are back in the linear separation case. To continue, let $T(p) = {N+p\choose p}$ be the number of monomials of degree less or equal than $p.$ For any listing $k\to \bn(k)$ of that set, denote by $c(k)$ the coefficient $c(\bn(k))$ and write the polynomial as 
\[
f(\bx) = \sum_{k=1}^{T(p)} c(k)\bx^{\bn(k)}.
\]
Under the same listing, we can think of all monomials of degree less or equal to $p$ as a vector $\bxi$ of length $T(p)$ with components $\xi(k)=\bx^{\bn(k)}$ so that 
\[
f(\bxi) = \sum_{k=1}^{T(p)}\xi(k)\,c(k).
\]
The notational analogy with the linear case should be clear by now. Form the matrices $\bG$ and $\bD$ of sizes $K_1\times T(p)$ and $K_2\times T(p),$ whose rows are, respectively, $\bg_i^{\bn(k)}$ for $i=1,\ldots,K_1,$ and $\bd_i^{\bn(k)}$ for $i=1,\ldots,K_2.$

As above, let us put 
\[
\bA_0 = {\bG \atopwithdelims[] -\bD},\qquad \bx={\bc^t \atopwithdelims() \bob},
\]
and then $\bA = [\bA_0, -\bbI_M],$ where, as above, $M=K_1+K_2$. Recall also that $\bob^t=(\bob_1^t,\bob_2^t).$

With all this, the separation by a polynomial surface $f(\bx)$ becomes exactly the same, except for the fact that we have yet to specify the set of constraints for the components of the vector $c(k): k=1,\ldots,T(p).$

\begin{problem}\label{P6}
Find $\bx^*\in\cK$ such that
\begin{equation}\label{prob6}
\bx^* = \arg\min\{\Psi(\bx)\mid \bx\in\cK,\;\bA\bx=\b0\}.
\end{equation}
\end{problem}

Here 
\[
\cK = [-e,e]^{T(p)} \times \bbR^M_{++},
\]
where $e>0$ is a parameter to be determined by the model builder.

\section{The entropy function}\label{sec:3}
We choose the function $\Psi(\bx):\cK\to\bbR$ in such a way that the box constraints are automatically satisfied. It is convenient to separate $\bx$ as a $N+M$-vector $(\bw,\bob)$. The function $\Psi(\bx)$ is defined as follows:
\begin{equation}\label{obj}
\Psi(\bx) \;=\; \sum_{j=1}^{N}\Biggl[\frac{w_j+1}{2}\ln\!\Bigl(\tfrac{w_j+1}{2}\Bigr) + \frac{1-w_j}{2}\ln\!\Bigl(\tfrac{1-w_j}{2}\Bigr)\Biggr] \;+\; \sum_{j=1}^{M}b_j(\ln b_j-1).
\end{equation}

The function $\Psi$ happens to be the Lagrange–Fenchel dual of the function \cite{BL}
\begin{equation}\label{mgf}
M(\btau,\bzeta) \;=\; \sum_{j=1}^{N} \ln\bigl(e^{ -\tau_j}+e^{\tau_j}\bigr)\;+\;\sum_{j=1}^M e^{\zeta_j},\qquad (\btau,\bzeta)\in\bbR^{\,N+M},
\end{equation}
which is the logarithm of the Laplace transform of the following measure:
\begin{equation}\label{remeas}
dQ(\bx) \;=\; \prod_{j=1}^N\bigl(\delta_{-1}(dw_j)+\delta_{1}(dw_j)\bigr)\;\prod_{j=1}^M\Bigl(\sum_{n=0}^\infty\frac{1}{n!}\,\delta_n(db_j)\Bigr)
\end{equation}
a measure that puts unit mass at each corner of $\cK$. The functions $\Psi$ and $M$ are related by:
\begin{equation}\label{FD1}
\begin{aligned}
\Psi(\bx)\;=\;\Psi(\bw,\bob) \;=&\;\sup\Bigl\{\langle\bw,\btau\rangle + \langle\bob,\bzeta\rangle \;-\; M(\btau,\bzeta)\;\Big|\;(\btau,\bzeta)\in\bbR^{\,N+M}\Bigr\},\quad \bx=(\bw,\bob)\in\cK,\\
M(\btau,\bzeta) \;=&\;\sup\Bigl\{ \langle\btau,\bw\rangle + \langle\bob,\bzeta\rangle \;-\; \Psi(\bx)\;\Big|\;\bx\in\cK\Bigr\},\quad (\btau,\bzeta)\in\bbR^{\,N+M}.
\end{aligned}
\end{equation}

It is simple to verify that both $M(\btau,\bzeta)$ and $\Psi(\bx)$ are convex. Furthermore, $\Psi$ is infinitely differentiable in the interior of $\cK$, and the following identity holds:
\begin{equation}\label{FD2}
\nabla M(\btau,\bzeta) \;=\; \bigl(\nabla \Psi\bigr)^{-1}(\btau,\bzeta),\quad\text{when}\;\;\nabla\Psi(\bx) = (\btau,\bzeta).
\end{equation}
That is, the gradients are inverse functions of each other. In particular, for every $(\btau,\bzeta)\in\bbR^{\,N+M}$, the equation $\bx = \nabla M(\btau,\bzeta)$ has a unique solution in the interior of $\cK$. Thus, for $\,1\le j\le N\,$:
\begin{align}\label{inv0}
w_j \;=\; \nabla_{\tau_j}M(\btau,\bzeta) &= \frac{-e^{-\tau_j} + e^{\tau_j}}{\,e^{-\tau_j} + e^{\tau_j}\,},\quad j=1,\ldots,N,\\
b_i \;=\;  \nabla_{\zeta_i}M(\btau,\bzeta) &= e^{\zeta_i},\quad i=1,\ldots,M.\nonumber
\end{align}

\section{Solving problem \ref{P4}}\label{sec:4}
Now we address the task of solving problem \ref{P4}, that is, to find $\bx^*\in\cK$ such that
\[
\bx^* \;=\; \arg\min\{\Psi(\bx)\mid \bx\in\cK,\;\bA\bx=\b0\},
\]
where $\Psi(\bx)$ is given by \eqref{obj}, namely
\[
\Psi(\bx) \;=\; \sum_{j=1}^{N}\Bigl[\tfrac{w_j+1}{2}\ln\!\bigl(\tfrac{w_j+1}{2}\bigr) + \tfrac{1-w_j}{2}\ln\!\bigl(\tfrac{1-w_j}{2}\bigr)\Bigr] \;+\; \sum_{j=1}^{M}b_j(\ln b_j - 1).
\]
For that we form the Lagrangian
\[
\cL(\bx,\blambda) \;=\; \Psi(\bx)\;-\;\langle\blambda,\bA^t\bx\rangle,
\]
and equate the derivatives with respect to $\bw$ and $\bob$ to zero to obtain:
\begin{gather}
\nabla_{\bw}\cL = \b0 
\;\;\Longrightarrow\;\;
\tfrac{1}{2}\ln\!\Bigl(\tfrac{1+w^*_j}{1-w^*_j}\Bigr) 
= (\bD^t\blambda^*)_j,\quad j=1,\ldots,N, \label{sol1}\\
\nabla_{\bob}\cL = \b0 
\;\;\Longrightarrow\;\;
-\lambda^*_i = \ln b^*_i,\quad i=1,\ldots,M. \label{sol2}
\end{gather}
After some simple computations, we obtain:
\begin{gather}
w^*_j \;=\; \frac{-e^{-(\bD^t\blambda^*)_j}+e^{(\bD^t\blambda^*)_j}}{\,e^{-(\bD^t\blambda^*)_j} + e^{(\bD^t\blambda^*)_j}\,},\quad j=1,\ldots,N,\label{repsol1.1}\\
b^*_i \;=\; e^{-\lambda^*_i},\quad i=1,\ldots,M.\label{repsol1.2}
\end{gather}

Recall that the first $K_1$ components of $\bob^*$ are those of $\bob_1^*$, and the next $K_2$ are those of $\bob_2^*$. To determine $\blambda^*$, note that 
\[
\cL(\bx^*,\blambda^*) = -M(\bA^t\blambda^*),
\]
and that the first-order condition for $\blambda^*$ to be a minimizer of $\blambda\mapsto M(\bA^t\blambda)$ are exactly \eqref{repsol1.1}–\eqref{repsol1.2}. (Note that $\bA^t\blambda^* \in \bbR^{\,N+M}$ so the first $N$ components are those of $\btau$ and the following $M$ components are those of $\bzeta$ in $M(\btau, \bzeta)$.)

Collecting the statements above we have a proof of the following result.

\begin{theorem}\label{T1}
The solution $\bx^*=(\bw^*,\bob^*)$ at which 
\[
\min\{\Psi(\bx)\mid \bx\in\cK,\;\bA\bx=\b0\}
\]
is reached is given by \eqref{repsol1.1}–\eqref{repsol1.2}, where the Lagrange multiplier $\blambda^*$ is the point at which the dual objective function $\blambda\mapsto M(\bA^t\blambda)$ reaches its minimum, with $M$ as in \eqref{mgf}.
\end{theorem}

Two comments are important at this point. First, the fact that $\blambda^*$ is the minimizer of $\blambda\mapsto M(\bA^t\blambda)$ makes its determination using standard spectral gradient methods much easier than solving \eqref{sol1}–\eqref{sol2} directly.

Second, if the equation satisfied by $\bx$ were $\bA\bx=\by$ with $\by\neq\b0$, then the dual Lagrangian would be $\langle\blambda,\by\rangle - M(\bA^t\blambda)$.

To conclude, once we have \eqref{repsol1.1} and \eqref{repsol1.2}, we can compute
\begin{equation}\label{eq:bounds}
\begin{aligned}
b_{+} \;=\;& \min\{\bob_1^*(i)\;:\; i=1,\ldots,K_1\} \;=\; \min\{\langle\bw^*,\bg_i\rangle \;:\; i=1,\ldots,K_1\},\\
b_{-} \;=\;& \min\{\bob_2^*(i)\;:\; i=1,\ldots,K_2\} \;=\; \min\{\langle\bw^*,-\bd_i\rangle \;:\; i=1,\ldots,K_2\}. 
\end{aligned}
\end{equation}
The equality in each line follows from the fact that $\bA\bx^*=\b0 \;\Leftrightarrow\; \bD\bw^*=\bob^*.$ With this, one can compute what \cite{CZ} call 
$
\Delta(\bw^*) \;\equiv\; b_{+} - b_{-},
$
which is a measure of the separation between the training data sets. The two hyperplanes 
\[
H_g \;=\; \{\bx\in\bbR^N\mid \langle\bw^*,\bx\rangle = b_{+}\}, 
\quad 
H_d \;=\; \{\bx\in\bbR^N\mid \langle\bw^*,\bx\rangle = b_{-}\},
\]
separate the data. For any new point $\bx \in \bbR^N$, if $\langle\bw^*,\bx\rangle \ge b_+$ or $\langle\bw^*,\bx\rangle \le b_{-}$, the point is correctly classified either as good or defective. However, if 
\[
b_{-} < \langle\bw^*,\bx\rangle < b_{+},
\]
the point is classified with a degree of uncertainty.

\subsection{The quality of the solution}
The fact that $\blambda^*$ is the minimizer of $\blambda\mapsto M(\bA^t\blambda)$ yields the following measure of the quality of the solution. Since $\bx^*$ must satisfy $\bA\bx^*=\b0$ to be considered a solution to Problem \ref{P4}, and bearing in mind that the first-order condition for $\blambda^*$ to be a minimizer is 
\[
\nabla_{\blambda}M(\bA^t\blambda^*) \;=\; \bA\,\bx^*(\blambda^*) \;=\; \b0,
\]
with $\bx^*(\blambda^*)$ given by \eqref{repsol1.1}–\eqref{repsol1.2}, one can stop the iterative numerical procedure to determine $\blambda^*$ when 
\[
\|\nabla_{\blambda}M(\bA^t\blambda^*)\| \;\le\; \varepsilon,
\]
(e.g.\ $\varepsilon=10^{-5}$). We then say that the data constraint is satisfied up to order $\varepsilon$.

\section{Experiments and Results}\label{sec:experiments}

To evaluate our findings, we compare the performance of the proposed Entropic Classifier (both linear and polynomial) against traditional classification models: Support Vector Machines (SVM), K-Nearest Neighbors (KNN), Logistic Regression, and Perceptron. The classifiers were tested across five different datasets to assess efficacy in various scenarios, including linearly and non-linearly separable data. 

\subsection{Datasets}

We considered five datasets generated and loaded from scikit-learn. Four of these datasets were synthetically generated (Blobs, Circles, Spirals, and Moons datasets) while the fifth dataset was real-world data (Breast Cancer dataset). The Blobs dataset represents two distinct linearly separable clusters. The Circles dataset consists of two concentric circles. The Spirals and Moons datasets are made of two interleaving spirals and two interleaving half circles, respectively. All synthetic datasets contained 2 features and 500 samples. The Breast Cancer dataset contained 30 features and 569 samples.

\subsection{Implementation Details}

All traditional classifiers (SVM, Logistic Regression, Perceptron, and KNN) were implemented using the scikit-learn library. For SVM and KNN, we performed hyperparameter optimization. The Entropic classifiers (linear and polynomial versions) were implemented based on the methodologies described in Sections \ref{sec:4} and \ref{sec:2}. We refer to our Entropic classifiers as Entropic-Linear (based on linear hyperplanes) and Entropic-Polynomial (based on polynomial surfaces as described in Section \ref{sec:2}). Performance metrics evaluated include Accuracy, Precision, Recall, F1-Score, and Confusion Matrices. We used a 70:30 train-test split for all models. For Entropic-Polynomial we performed a grid search to determine the best degree for each dataset (two, three, and four were considered, as higher degrees risk overfitting).

\subsection{Results}

\subsubsection{Blobs Dataset}

\begin{table}[ht]
    \centering
    \caption{Performance Comparison on Blobs Dataset}
    \label{tab:blobs_results}
    \begin{tabular}{lcccc}
        \toprule
        \textbf{Method} & \textbf{Accuracy} & \textbf{Precision} & \textbf{Recall} & \textbf{F1-Score} \\
        \midrule
        Entropic-Linear & 1.00 & 1.00 & 1.00 & 1.00 \\
        SVM & 1.00 & 1.00 & 1.00 & 1.00 \\
        KNN & 1.00 & 1.00 & 1.00 & 1.00 \\
        Logistic Regression & 1.00 & 1.00 & 1.00 & 1.00 \\
        Perceptron & 1.00 & 1.00 & 1.00 & 1.00 \\
        Entropic-Polynomial (Degree 2) & 1.00 & 1.00 & 1.00 & 1.00 \\
        \bottomrule
    \end{tabular}
\end{table}

\begin{table}[ht]
    \centering
    \caption{Confusion Matrices for Blobs Dataset}
    \label{tab:confusion_blobs}
    \begin{tabular}{lcc|cc}
        \toprule
        \multirow{2}{*}{\textbf{Classifier}} & \multicolumn{2}{c|}{\textbf{Predicted: Class 0}} & \multicolumn{2}{c}{\textbf{Predicted: Class 1}} \\
         & \textbf{True 0} & \textbf{True 1} & \textbf{True 0} & \textbf{True 1} \\
        \midrule
        Entropic-Linear & 70 & 0 & 0 & 80 \\
        SVM & 70 & 0 & 0 & 80 \\
        KNN & 70 & 0 & 0 & 80 \\
        Logistic Regression & 70 & 0 & 0 & 80 \\
        Perceptron & 70 & 0 & 0 & 80 \\
        Entropic-Polynomial (Degree 2) & 70 & 0 & 0 & 80 \\
        \bottomrule
    \end{tabular}
\end{table}

\paragraph{Discussion}
For the Blobs dataset (Tables \ref{tab:blobs_results} and \ref{tab:confusion_blobs}), all classifiers—traditional and entropic—achieved perfect performance, as expected for perfectly linearly separable data.

\subsubsection{Circles Dataset}

\begin{table}[ht]
    \centering
    \caption{Performance Comparison on Circles Dataset}
    \label{tab:circles_results}
    \begin{tabular}{lcccc}
        \toprule
        \textbf{Method} & \textbf{Accuracy} & \textbf{Precision} & \textbf{Recall} & \textbf{F1-Score} \\
        \midrule
        Entropic-Linear & 0.46 & 0.00 & 0.00 & 0.00 \\
        SVM & 0.53 & 0.52 & 0.64 & 0.57 \\
        KNN & 0.97 & 1.00 & 0.93 & 0.97 \\
        Logistic Regression & 0.49 & 0.49 & 0.48 & 0.48 \\
        Perceptron & 0.57 & 0.55 & 0.76 & 0.64 \\
        Entropic-Polynomial (Degree 4) & 0.99 & 0.99 & 0.99 & 0.99 \\
        \bottomrule
    \end{tabular}
\end{table}

\begin{table}[ht]
    \centering
    \caption{Confusion Matrices for Circles Dataset}
    \label{tab:confusion_circles}
    \begin{tabular}{lcc|cc}
        \toprule
        \multirow{2}{*}{\textbf{Classifier}} & \multicolumn{2}{c|}{\textbf{Predicted: Class 0}} & \multicolumn{2}{c}{\textbf{Predicted: Class 1}} \\
         & \textbf{True 0} & \textbf{True 1} & \textbf{True 0} & \textbf{True 1} \\
        \midrule
        Entropic-Linear & 69 & 6 & 75 & 0 \\
        SVM & 31 & 44 & 27 & 48 \\
        KNN & 75 & 0 & 5 & 70 \\
        Logistic Regression & 37 & 38 & 39 & 36 \\
        Perceptron & 28 & 47 & 18 & 57 \\
        Entropic-Polynomial (Degree 4) & 74 & 1 & 1 & 74 \\
        \bottomrule
    \end{tabular}
\end{table}

\paragraph{Discussion}
For the Circles dataset (Tables \ref{tab:circles_results} and \ref{tab:confusion_circles}), all linear classifiers performed poorly (as expected for non-linearly separable data). The Entropic-Linear and Logistic Regression models performed worse than random guessing; SVM and Perceptron were only marginally better. KNN achieved very high accuracy, and Entropic-Polynomial (degree 4) achieved nearly perfect performance, misclassifying just two points.

\subsubsection{Spiral Dataset}

\begin{table}[ht]
    \centering
    \caption{Performance Comparison on Spiral Dataset}
    \label{tab:spiral_results}
    \begin{tabular}{lcccc}
        \toprule
        \textbf{Method} & \textbf{Accuracy} & \textbf{Precision} & \textbf{Recall} & \textbf{F1-Score} \\
        \midrule
        Entropic-Linear & 0.72 & 1.00 & 0.44 & 0.61 \\
        SVM (GridSearch) & 0.74 & 0.75 & 0.72 & 0.73 \\
        KNN (GridSearch) & 1.00 & 1.00 & 1.00 & 1.00 \\
        Logistic Regression & 0.74 & 0.75 & 0.73 & 0.74 \\
        Perceptron & 0.31 & 0.31 & 0.31 & 0.31 \\
        Entropic-Polynomial (Degree 3) & 0.99 & 1.00 & 0.99 & 0.99 \\
        \bottomrule
    \end{tabular}
\end{table}

\begin{table}[ht]
    \centering
    \caption{Confusion Matrices for Spiral Dataset}
    \label{tab:confusion_spiral}
    \begin{tabular}{lcc|cc}
        \toprule
        \multirow{2}{*}{\textbf{Classifier}} & \multicolumn{2}{c|}{\textbf{Predicted: Class 0}} & \multicolumn{2}{c}{\textbf{Predicted: Class 1}} \\
         & \textbf{True 0} & \textbf{True 1} & \textbf{True 0} & \textbf{True 1} \\
        \midrule
        Entropic-Linear & 150 & 0 & 84 & 66 \\
        SVM & 114 & 36 & 42 & 108 \\
        KNN & 150 & 0 & 0 & 150 \\
        Logistic Regression & 113 & 37 & 41 & 109 \\
        Perceptron & 45 & 105 & 103 & 47 \\
        Entropic-Polynomial (Degree 3) & 150 & 0 & 2 & 148 \\
        \bottomrule
    \end{tabular}
\end{table}

\paragraph{Discussion}
For the Spiral dataset (Tables \ref{tab:spiral_results} and \ref{tab:confusion_spiral}), Entropic-Linear (like the other linear classifiers) achieved moderate accuracy but low recall; it classified all Class 0 correctly but struggled with Class 1. Traditional linear classifiers had similar behavior. Entropic-Polynomial (degree 3) nearly matched KNN, misclassifying only two points.

\subsubsection{Moons Dataset}

\begin{table}[ht]
    \centering
    \caption{Performance Comparison on Moons Dataset}
    \label{tab:moons_results}
    \begin{tabular}{lcccc}
        \toprule
        \textbf{Method} & \textbf{Accuracy} & \textbf{Precision} & \textbf{Recall} & \textbf{F1-Score} \\
        \midrule
        Entropic-Linear & 0.80 & 1.00 & 0.60 & 0.75 \\
        SVM (GridSearch) & 0.88 & 0.88 & 0.88 & 0.88 \\
        KNN (GridSearch) & 1.00 & 1.00 & 1.00 & 1.00 \\
        Logistic Regression & 0.88 & 0.88 & 0.88 & 0.88 \\
        Perceptron & 0.87 & 0.88 & 0.87 & 0.87 \\
        Entropic-Polynomial (Degree 3) & 0.99 & 1.00 & 0.97 & 0.99 \\
        \bottomrule
    \end{tabular}
\end{table}

\begin{table}[ht]
    \centering
    \caption{Confusion Matrices for Moons Dataset}
    \label{tab:confusion_moons}
    \begin{tabular}{lcc|cc}
        \toprule
        \multirow{2}{*}{\textbf{Classifier}} & \multicolumn{2}{c|}{\textbf{Predicted: Class 0}} & \multicolumn{2}{c}{\textbf{Predicted: Class 1}} \\
         & \textbf{True 0} & \textbf{True 1} & \textbf{True 0} & \textbf{True 1} \\
        \midrule
        Entropic-Linear & 75 & 0 & 30 & 45 \\
        SVM & 66 & 9 & 9 & 66 \\
        KNN & 75 & 0 & 0 & 75 \\
        Logistic Regression & 66 & 9 & 9 & 66 \\
        Perceptron & 66 & 9 & 10 & 65 \\
        Entropic-Polynomial (Degree 3) & 75 & 0 & 2 & 73 \\
        \bottomrule
    \end{tabular}
\end{table}

\paragraph{Discussion}
For the Moons dataset (Tables \ref{tab:moons_results} and \ref{tab:confusion_moons}), Entropic-Linear performed reasonably but had lower recall. Traditional linear classifiers were better overall. Entropic-Polynomial (degree 3) again closely matched KNN’s perfect boundary.

\subsubsection{Breast Cancer}

\begin{table}[ht]
    \centering
    \caption{Performance Comparison on Breast Cancer Dataset}
    \label{tab:breast_cancer_results}
    \begin{tabular}{lcccc}
        \toprule
        \textbf{Method} & \textbf{Accuracy} & \textbf{Precision} & \textbf{Recall} & \textbf{F1-Score} \\
        \midrule
        Entropic-Linear & 0.94 & 0.99 & 0.91 & 0.95 \\
        SVM & 0.98 & 0.98 & 0.99 & 0.99 \\
        KNN & 0.96 & 0.96 & 0.97 & 0.97 \\
        Logistic Regression & 0.98 & 0.99 & 0.98 & 0.99 \\
        Perceptron & 0.98 & 0.99 & 0.98 & 0.99 \\
        Entropic-Polynomial (Degree 2) & 0.92 & 0.94 & 0.94 & 0.94 \\
        \bottomrule
    \end{tabular}
\end{table}

\begin{table}[ht]
    \centering
    \caption{Confusion Matrices for Breast Cancer Dataset}
    \label{tab:confusion_breast_cancer}
    \begin{tabular}{lcc|cc}
        \toprule
        \multirow{2}{*}{\textbf{Classifier}} & \multicolumn{2}{c|}{\textbf{Predicted: Class 0}} & \multicolumn{2}{c}{\textbf{Predicted: Class 1}} \\
         & \textbf{True 0} & \textbf{True 1} & \textbf{True 0} & \textbf{True 1} \\
        \midrule
        Entropic-Linear & 62 & 1 & 10 & 98 \\
        SVM & 61 & 2 & 1 & 107 \\
        KNN & 58 & 5 & 3 & 105 \\
        Logistic Regression & 62 & 1 & 2 & 106 \\
        Perceptron & 62 & 1 & 2 & 106 \\
        Entropic-Polynomial (Degree 2) & 56 & 7 & 7 & 101 \\
        \bottomrule
    \end{tabular}
\end{table}

\paragraph{Discussion}
For the Breast Cancer dataset (Tables \ref{tab:breast_cancer_results} and \ref{tab:confusion_breast_cancer}), all models performed well. Notably, Entropic-Linear—despite struggling on non-linearly separable synthetic datasets—performed strongly in this real-world dataset. Entropic-Polynomial (degree 2) also did well, as the 30-feature space permitted a low-degree polynomial to capture the decision boundary without severe overfitting.

\subsection{Visualization of Decision Boundaries}

\begin{figure}[ht]
    \centering
    \includegraphics[width=\textwidth]{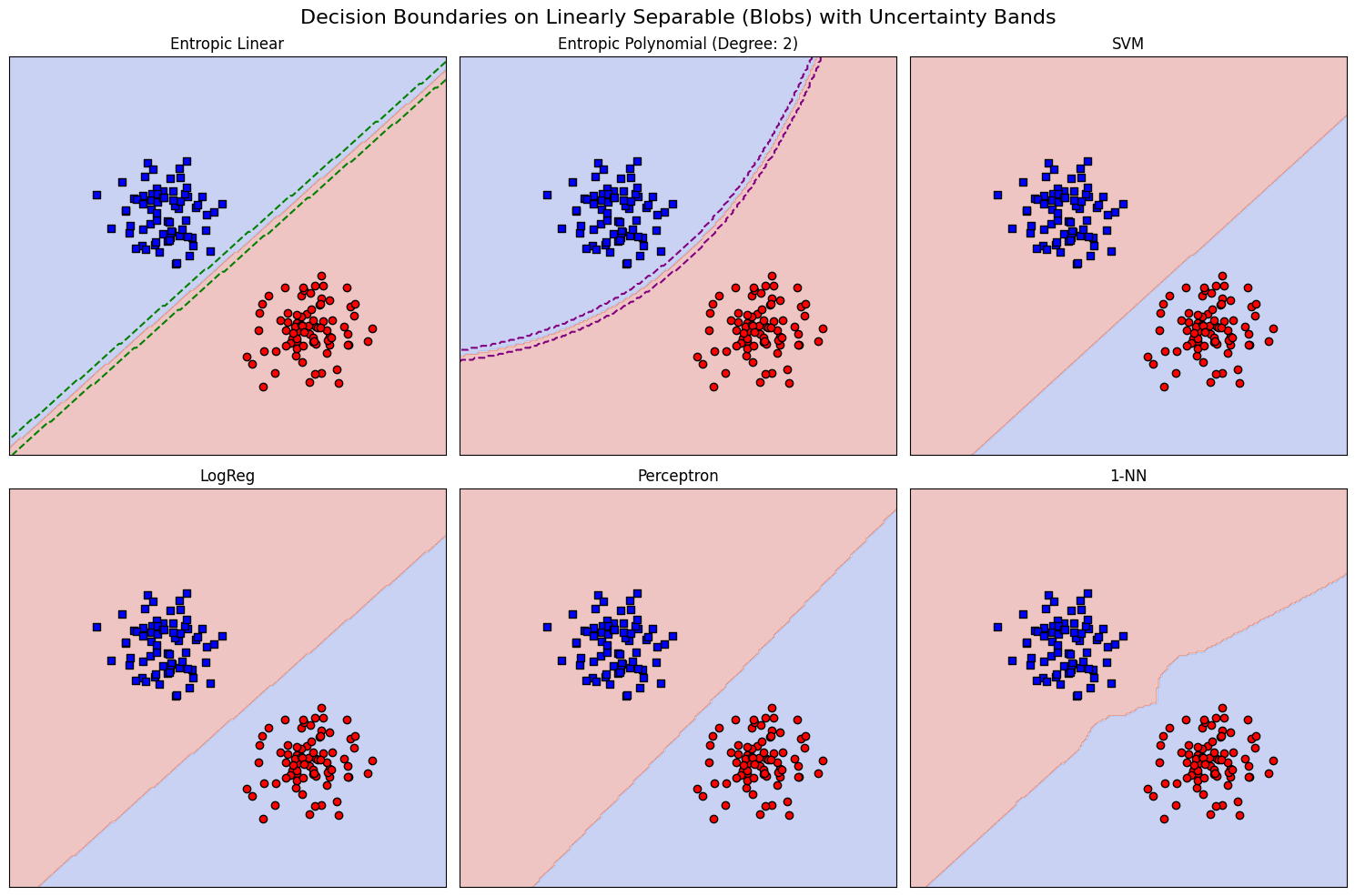}
    \caption{Decision Boundaries on Blobs Dataset with Uncertainty Bands}
    \label{fig:blobs}
\end{figure}

\begin{figure}[ht]
    \centering
    \includegraphics[width=\textwidth]{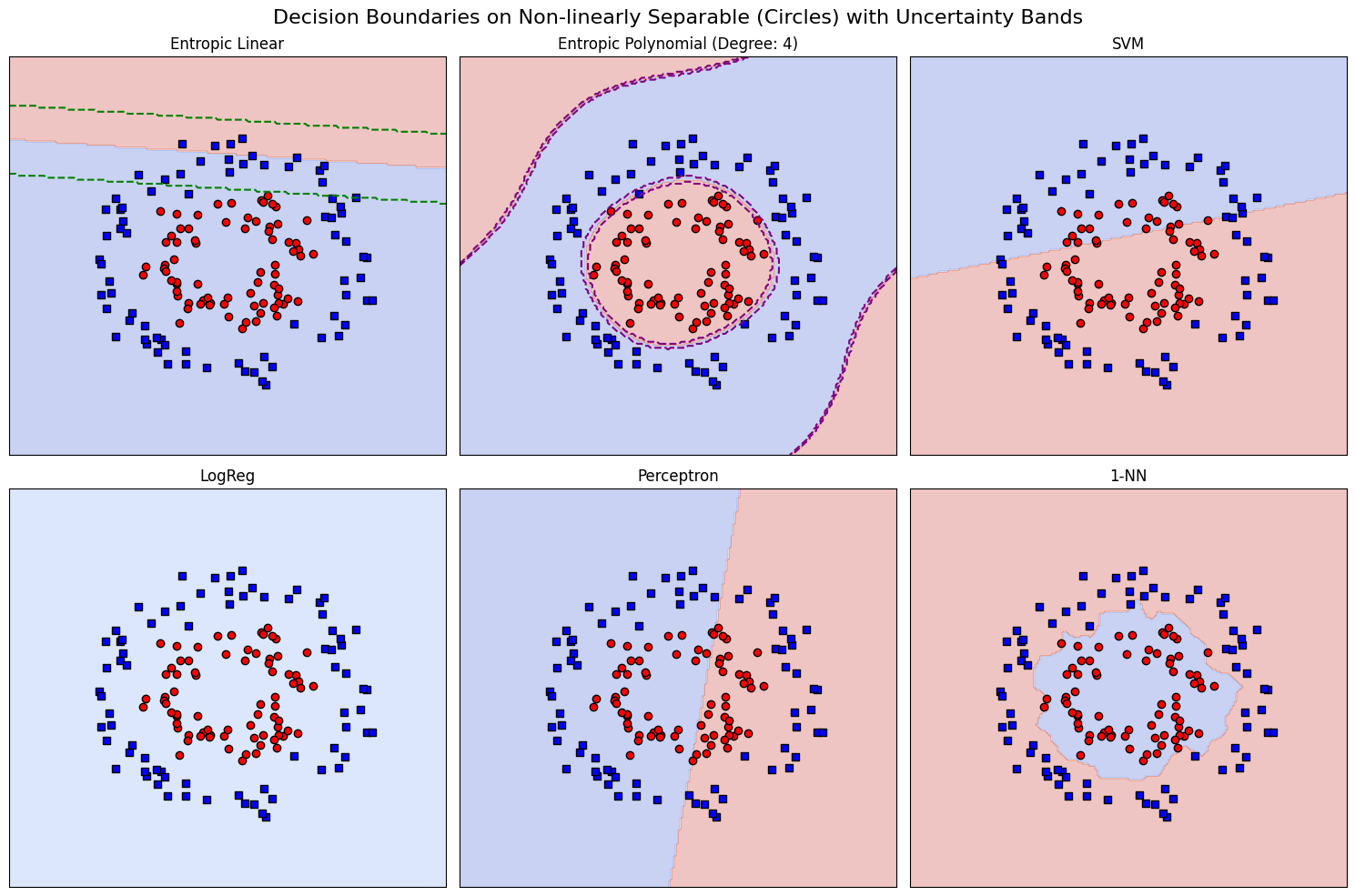}
    \caption{Decision Boundaries on Circles Dataset with Uncertainty Bands}
    \label{fig:circles}
\end{figure}

\begin{figure}[ht]
    \centering
    \includegraphics[width=\textwidth]{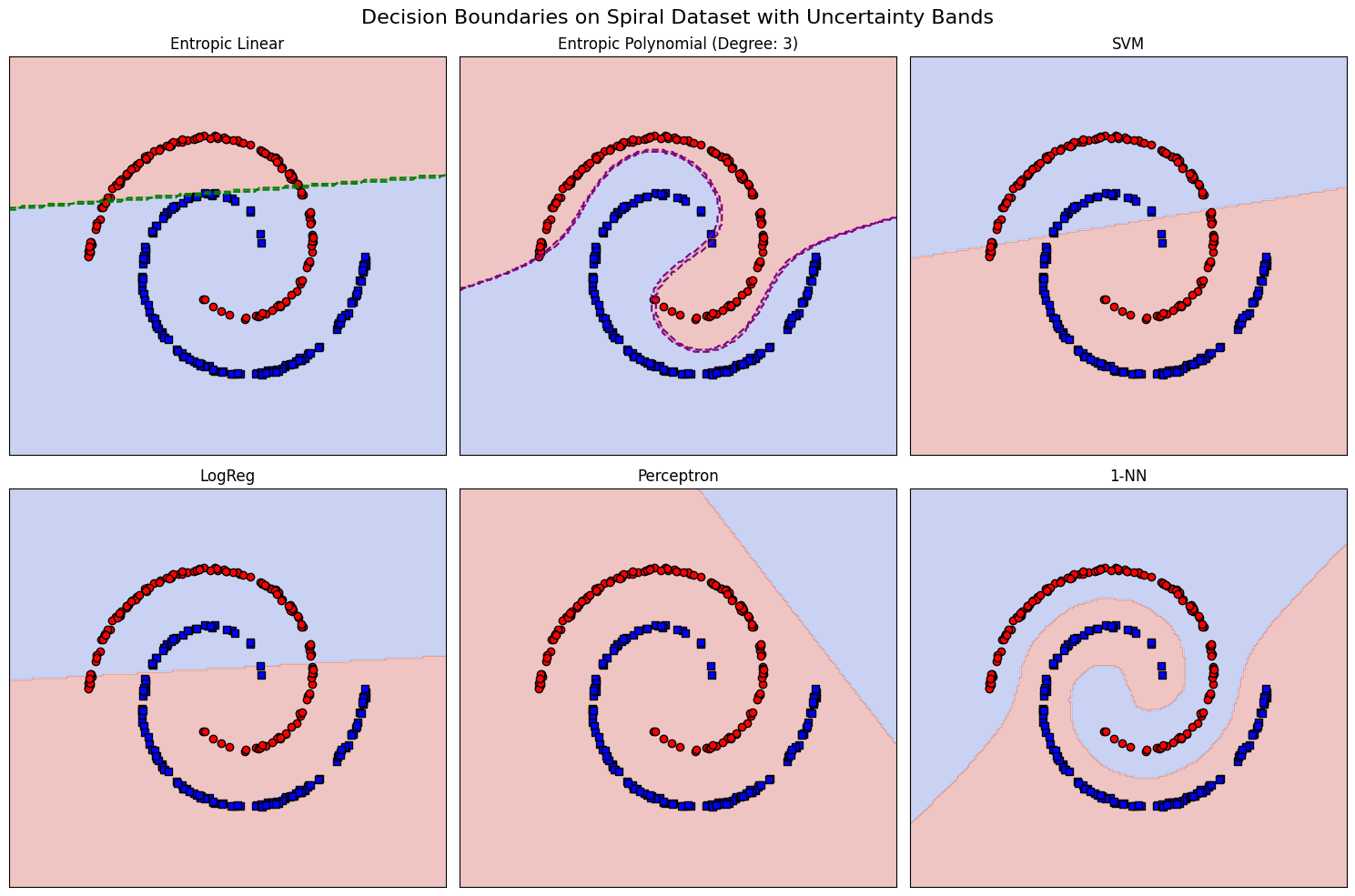}
    \caption{Decision Boundaries on Spiral Dataset with Uncertainty Bands}
    \label{fig:spiral}
\end{figure}

\begin{figure}[ht]
    \centering
    \includegraphics[width=\textwidth]{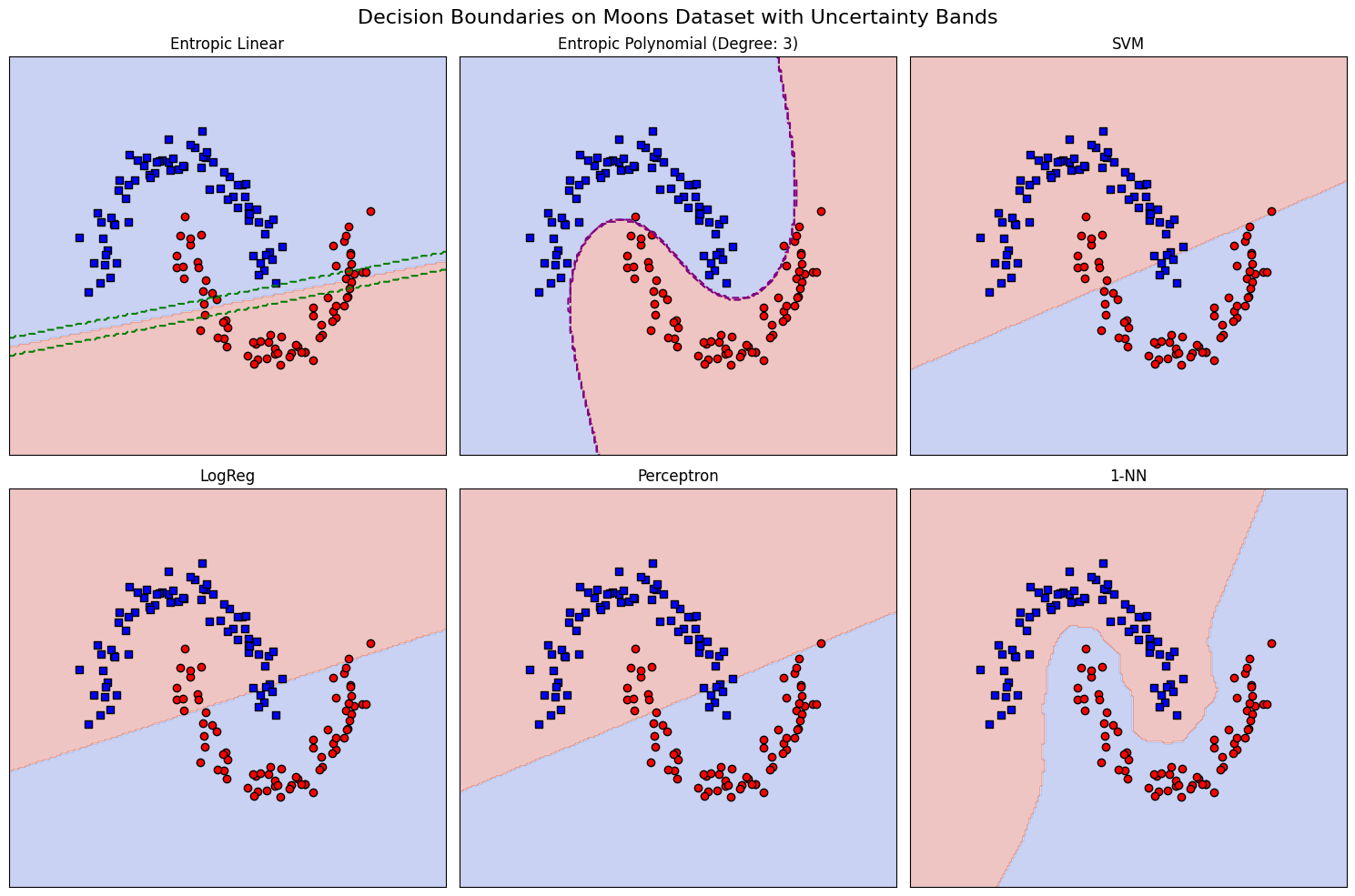}
    \caption{Decision Boundaries on Moons Dataset with Uncertainty Bands}
    \label{fig:moons}
\end{figure}

Figures \ref{fig:blobs}, \ref{fig:circles}, \ref{fig:spiral}, and \ref{fig:moons} show decision boundaries of each classifier for the Blobs, Circles, Spiral, and Moons datasets, respectively. For the Entropic classifiers, the uncertainty bands (between $H_d$ and $H_g$ as in \eqref{eq:bounds}) are included as dashed lines.

\subsection{Summary}
As a summary, the Blobs dataset (linearly separable) led all models—entropic and traditional—to perfect metrics. In the non-linearly separable datasets (Circles, Spiral, and Moons), linear models (both traditional and entropic) struggled to capture curved boundaries, while KNN and Entropic-Polynomial easily handled the nonlinearity. The figures confirm this: linear classifiers draw straight lines, whereas KNN and Entropic-Polynomial produce boundaries that mirror the data’s shape. The dashed lines in the entropic plots are the uncertainty bands (where classification is with uncertainty). Finally, on the real-world Breast Cancer dataset, both Entropic-Linear and Entropic-Polynomial performed competitively with traditional methods.

\section*{Statements and Declarations}
\subsection*{Competing Interests}
The authors report there are no competing interests to declare.

\subsection*{Funding}
No funding was received for conducting this study.

\subsection*{Data and Code Availability}
The data and Python code to reproduce the results in this work are available on GitHub: 
\url{https://github.com/argimiroa/EntropicClassifier.git}

\end{document}